\documentclass[conference]{IEEEtran}
\IEEEoverridecommandlockouts
\usepackage{cite}
\usepackage{amsmath,amssymb,amsfonts}
\usepackage{algorithmic}
\usepackage{graphicx}
\usepackage{textcomp}
\usepackage{xcolor}
\usepackage{caption}
\usepackage{subcaption}
\def\BibTeX{{\rm B\kern-.05em{\sc i\kern-.025em b}\kern-.08em
    T\kern-.1667em\lower.7ex\hbox{E}\kern-.125emX}}
\begin{document}

\title{Offline Reinforcement Learning for Mobile Notifications}

\author{
\IEEEauthorblockN{Yiping Yuan}
\IEEEauthorblockA{\textit{LinkedIn Corporation} \\
Mountain View, CA, USA \\
ypyuan@linkedin.com}
\and
\IEEEauthorblockN{Ajith Muralidharan}
\IEEEauthorblockA{\textit{LinkedIn Corporation} \\
Mountain View, CA, USA \\
amuralidharan@linkedin.com}
\and
\IEEEauthorblockN{Preetam Nandy}
\IEEEauthorblockA{\textit{LinkedIn Corporation} \\
Mountain View, CA, USA \\
pnandy@linkedin.com}
\and
\IEEEauthorblockN{Miao Cheng}
\IEEEauthorblockA{\textit{LinkedIn Corporation} \\
Mountain View, CA, USA \\
miacheng@linkedin.com}
\and
\IEEEauthorblockN{ Prakruthi Prabhakar}
\IEEEauthorblockA{\textit{LinkedIn Corporation} \\
Mountain View, CA, USA \\
paprabhakar@linkedin.com}

}

\maketitle

\begin{abstract}
Mobile notification systems have taken a major role in driving and maintaining user engagement for online platforms. They are interesting recommender systems to machine learning practitioners with more sequential and long-term feedback considerations. Most machine learning applications in notification systems are built around response-prediction models, trying to attribute both short-term impact and long-term impact to a notification decision. However, a user's experience depends on a sequence of notifications and attributing impact to a single notification is not always accurate, if not impossible. In this paper, we argue that reinforcement learning is a better framework for notification systems in terms of performance and iteration speed. We propose an offline reinforcement learning framework to optimize sequential notification decisions for driving user engagement. We describe a state-marginalized importance sampling policy evaluation approach, which can be used to evaluate the policy offline and tune learning hyperparameters. Through simulations that approximate the notifications ecosystem, we demonstrate the performance and benefits of the offline evaluation approach as a part of the reinforcement learning modeling approach. Finally, we collect data through online exploration in the production system, train an offline Double Deep Q-Network and launch a successful policy online. We also discuss the practical considerations and results obtained by deploying these policies for a large-scale recommendation system use-case.
\end{abstract}


\section{Introduction}
As online services and applications provide more and more content and functionality, communications with users are increasingly crucial for them to keep users informed and engaged. Mobile notifications are a major channel that services use to highlight important and timely content to the users. With the right content at the right time, notifications can inform users of important activity and bring more value to users. Since users have limited attention, notifications can help remind users of important values that they would need to be aware of to increase engagement with the platform. There are mainly two categories of responses from sending a notification:  content engagement responses (e.g., clicks, dismisses) and site engagement responses (e.g., user visits, notification disables). Typical recommender systems usually care more about the content engagement responses, while for notification systems, site engagement responses are as important if not more.  Unlike content engagement responses, user engagement may not be attributed to a single notification, but rather a sequence of notifications, presenting an attribution challenge for modeling. Another challenge for modeling site engagement responses is that the short-term (within a few hours) impact and long-term (over a week or longer) impact may diverge. Although  intrusive and frequent notifications can bring users back to site, they could create notification fatigue or cause notification disablement, which hurts user engagement in the long run \cite{pielot2014situ, okoshi2019real}.  These unique challenges give rise to an interesting application area to machine learning practitioners with more sequential and long-term considerations.

 Most notification systems \cite{gupta2016email,gupta2017optimizing,yuan2019state,gao2018near,zhao2018notification} are built around response prediction models. To overcome the attribution challenge of the site engagement responses,
 a state-transition model is proposed to predict the additional user visits attributed to a single notification in \cite{yuan2019state}. The volume optimization framework proposed in \cite{zhao2018notification} attributes site engagement responses to a weekly notification count rather than to a single notification. These response predictions are then compared with optimal thresholds from online or offline threshold search based on  multi-objective optimizations \cite{agarwal2011click}. While such systems have demonstrated good empirical performance over CTR-based systems, they could  be sub-optimal in decision making. First, the attribution is still approximate and cannot fully capture the sequential impact. For example, the volume optimization framework in \cite{zhao2018notification} assumes the site engagement response only depends on the volume sent to a user without considering the spacing of notification deliveries under the same volume. Secondly, the online or offline threshold tuning is often heuristic, and may not achieve the optimality defined by the multi-objective optimization. Practically, response prediction model improvement (e.g., in terms of offline AUC) may not necessarily lead to better online performance when couple with the threshold tuning. While there are efforts to automate and optimize this tuning \cite{talosTutorial}, it could slow down model iteration by weeks from trained response models to a fully deployed system. And the model iteration speed is a very important consideration to  real world systems that need to be improved and updated constantly.

In comparison, reinforcement learning is a principled approach to optimize for a sequence of well-coordinated notification decisions with respect to the defined objective. The attribution challenge comes down to the definition of the rewards. If rewards from the environment are defined properly, the aggregated rewards (the total return) will be consistent with the business objective. We explain our reward definition in Section \ref{sec:mdpSpacing}. Reinforcement learning is  a superior framework to emphasize long-term impact with the aggregated rewards over a long or infinite horizon. Moreover, the offline reinforcement learning framework we propose comes with efficient offline policy evaluation, which provide consistency between offline evaluation and online performance. It could also avoid costly online tuning and speed up model iteration as described in Section \ref{sec:offlineEvaluation}  and Section \ref{sec:experiments}.

Applying reinforcement learning to a large-scale online system faces several challenges. Online reinforcement learning training with online exploration may not be feasible due to high infrastructure costs, unknown time to converge, and unbounded risks of deteriorating user experience.  The fact that a large proportion of reinforcement learning algorithms and research are more focused towards online learning
paradigm is also one of the biggest obstacles to their widespread adoption\cite{levine2020offline}. Alternatively, offline reinforcement learning  \cite{lange2012batch,agarwal2019striving, chen2019top,ie2019reinforcement,fujimoto2019off,levine2020offline, kumar2020conservative} has started to draw more research attention in recent years due to its well-controlled risk and a smoother fit into existing machine learning infrastructure. There are theoretical and practical challenges in efficient offline policy learning, and accurate offline policy evaluation \cite{mahmood2014weighted,jiang2016doubly,thomas2016data,xie2019towards} due to the notorious Deadly Triad problem (i.e., the problem of instability and divergence
arising when combining function approximation, bootstrapping and offline training) \cite{sutton1998reinforcement,van2018deep}. Compared with classical control problems, the low signal-to-noise ratio and potential non-linearity in user behavior require thoughtful Markov Decision Process (MDP) formulation, adequate function approximation, and exploratory behavior policy design to learn effective policies offline. Additionally, it is important to have a reliable offline off-policy evaluation algorithm to ensure safe and efficient policy iterations \cite{xie2019towards}. 

In this paper, we propose an offline reinforcement learning approach to optimize for site engagement in notification systems. We summarize our contribution
\begin{itemize}
    \item We formulate a MDP to model the notification system and specify an offline learning approach based on the (Double) Deep Q-Network. 
    \item We propose a state-marginalized importance sampling algorithm for offline evaluation to reduce the high variance of the existing importance sampling based algorithms. 
    \item We evaluate our approach using a simplified simulation setup that helps mimic the online process we  optimize. We use this simulation environment to validate and benchmark offline evaluation methods.
    \item We present a real-world fully-deployed application to demonstrate how such a reinforcement learning paradigm can improve site user engagement and achieve better performance than the supervised approach.
\end{itemize}

The rest of this paper is organized as follows. Section \ref{sec:relatedWork} reviews related work. In Section \ref{sec:dto}, we introduce the problem of notification delivery time optimization and its Markov Decision Process formulation. Section \ref{sec:methodology} introduces the underlying methodology for offline training, offline evaluation and how we build up a simulation environment to mimic the real-world application. Section \ref{sec:experiments} carries out both simulated and real-world experiments to demonstrate how the proposed framework works. Finally, Section \ref{sec:conclusion} concludes this work and discusses our future work.

\section{Related Work}
\label{sec:relatedWork}
Early work in \cite{gupta2016email,gupta2017optimizing} proposed a volume optimization framework based on supervised model predictions. The framework was originally designed for emails and was later extended to notification applications \cite{gao2018near,zhao2018notification} with more considerations on real-time relevance and machine learning infrastructures. Other related work  \cite{mehrotra2015designing, pielot2014didn, pielot2017beyond} focused on improving the prediction with supervised learning. A survival-based state-transition model \cite{yuan2019state} was proposed to drive user engagement through mobile notifications with heuristic global trade-offs between short-term and long-term. A bandit-based solution \cite{wu2017returning} was proposed to improve long-term user engagement in a recommender system. These approaches worked well in practice but could be suboptimal in sequential decision making. We argue in this paper that a reinforcement learning framework is a better fit to notification systems.

In terms of notification fatigue and unnecessary interruptions, several notification systems \cite{okoshi2015attelia,okoshi2015reducing, pejovic2014interruptme, pielot2015attention} have been proposed for detecting opportune timings for delivery of notifications, leveraging various types of sensing and machine learning technologies. Such an adaptive notification system was evaluated in the product environment and showed impressive click-through rate increase when powered by supervised machine learning \cite{okoshi2017attention, okoshi2019real}. We introduce a similar production system at LinkedIn in Section \ref{sec:notificationSpacing}, which we call it ``Notification Spacing System''. A state-of-art  survival model \cite{yuan2019state} is used in this system as the supervised benchmark for our proposed reinforcement learning. This survival model was the best performing production model at LinkedIn before we introduce this work. In this paper, we illustrate how reinforcement learning can be applied to such large-scale production systems and is compared against the supervised benchmark.

Recently, Chen et al. \cite{chen2019top} applied a Policy Gradient learning in YouTube recommender system with Off-policy correction for offline reinforcement learning. Ie et al. \cite{ie2019reinforcement} proposed an offline reinforcement learning for slate-based recommender systems, where the large action space can become intractable for many reinforcement learning algorithms. Zou et al. \cite{zou2019reinforcement} designed an offline framework to learn a Q-network and a separate S-network from a simulated environment to assist the Q-network. In comparison, our simulated environment is only used for validation with ground truth, and the deployed policy takes no information from the simulated environment to avoid unknown bias.  While there are a lot of efforts to apply reinforcement learning to real-world applications \cite{zhao2018recommendations,hughes2019generating,wang2020incremental}, driving long-term engagement through notification systems 
 presents its unique challenges and opportunities for reinforcement learning due to its sequential planning and short-term long-term trade-off.

Offline evaluation is another crucial research area for real-world applications. Various importance sampling methods \cite{thomas2016data,mahmood2014weighted,jiang2016doubly,swaminathan2016off} have been applied to correct the mismatch in
the distributions under the behavior policy and evaluated policy. While these importance sampling based estimators are either unbiased or have little bias, the variances of them tend to be high for a long sequence.
A marginalized importance sampling \cite{xie2019towards} was recently proposed to reduce variances for long-horizon MDP environments, with extensive theoretical and empirical studies. We choose this method for the offline evaluation and  apply some practical modifications, namely a novel dimension reduction technique and discretization.

\section{Notification delivery time optimization}
\label{sec:dto}

\begin{figure}[h]
\center
\includegraphics[width=0.6\linewidth]{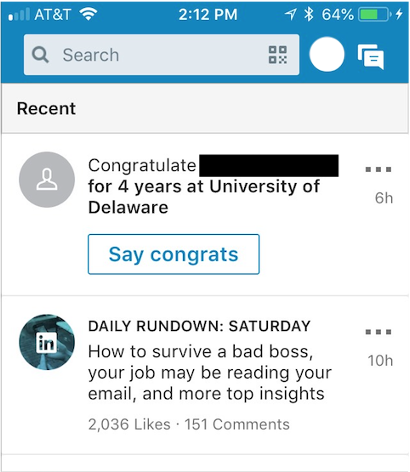}
\caption{\label{fig:example-notification} An example of time-insensitive notifications } 
\end{figure}

Online platforms use mobile notifications to communicate timely, important, and actionable content to users. Some notifications, due to their nature, user expectation, or product constraints, need to be delivered in near real-time. Examples of these include content notifications described in \cite{gao2018near} and notifications about messages sent to users. There are other notifications, which are not time-sensitive, and they would be relevant if they are delivered within a predefined time window. Examples of these notifications include events from your network, such as your colleague's work anniversary and birthday, or aggregate notifications about activities that you may be interested in. Figure \ref{fig:example-notification} gives such an example of work anniversary notification. Such time-insensitive notifications provide more opportunities for online platforms to optimize for site engagement through delivery time optimization. In this paper, we focus our discussions on applying reinforcement learning to such time-insensitive notifications to determine the best delivery times towards long-term engagement. We describe such a notification interaction environment through the notification spacing system at LinkedIn. This system ensures that users do not receive all the notifications at the same time but receive them over the course of multiple days or weeks, providing users a holistic and engaging experience over time.

\subsection{Notification Spacing System}
\label{sec:notificationSpacing}

\begin{figure}[h]
\center
\includegraphics[width=\linewidth]{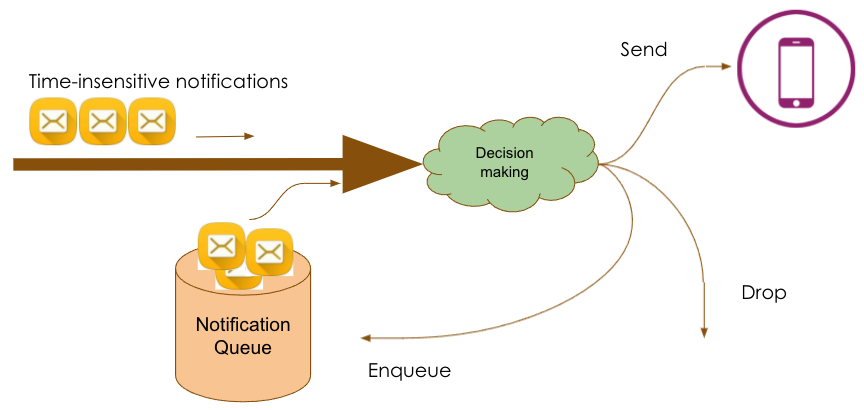}
\caption{\label{fig:spacing-system} Illustration of the notification spacing system } 
\end{figure}

The notification spacing system in Figure \ref{fig:spacing-system} consists of a queuing system (one for each user), into which time-insensitive notification candidates for the user are queued. At fixed time intervals, we choose whether to send the top-ranked notification in the queue. Additionally, at each such time step, there may be notifications that would expire after that time. The notification spacing system also determines which of these expiring notifications to send to the user and which ones to drop.

We now briefly describe the baseline policy in this system, which is based on the work presented in \cite{yuan2019state} (Section 5.2). This is the supervised approach that delivers the best empirical performance against other supervised approaches at LinkedIn. An Accelerated Failure Time survival model is trained using the user interaction data. The model is then used to predict 1) the probability of a user's visit within the next $T$ time after a notification delivery, denoted as $P_T(\text{visit}|\text{send})$; 2) the probability of a user's organic visit within the next $T$ time without a notification delivery, denoted as $P_T(\text{visit}|\text{not send})$. This baseline model chooses to send the top ranked notification if 
\begin{equation}
     \frac{P_T(\text{visit}|\text{send})-P_T(\text{visit}|\text{not send})}{P_T(\text{visit}|\text{not send})} > \tau ,
\end{equation}
where  \(\tau \) is the threshold to heuristically  control the trade-off between short-term and long-term rewards. $P_T(\text{visit}|\text{send})-P_T(\text{visit}|\text{not send})$ is the uplift estimate of the short-term impact on the user engagement by a notification delivery. The extra denominator $P_T(\text{visit}|\text{not send})$ helps to normalize with respect to a user's activity level. The policy makes a send decision when the short-term uplift exceeding the threshold $\tau$.  Note that \(\tau  = 0 \) leads to a greedy action, that is, as long as the short-term uplift is positive, a notification will be delivered to a user, which will almost surely result in annoying notification experience and hurt the long-term site engagement. A policy with a larger $\tau$ will withhold some notifications in the hope that there will be more opportune moment in the future. One drawback of this approach is that $\tau$ is hyper-parameter outside of the supervised learning, thus cannot be learned with the model training. Instead, $\tau$ is usually tuned through grid search using online A/B test. This is the case for supervised approaches for notification systems in general, since there is often a gap between the model's prediction and the optimal notification decision. 

\subsection{Markov Decision Process for Notification Spacing}
\label{sec:mdpSpacing}
In this section, we formulate the notification delivery time optimization in the notification spacing system as a Markov Decision Process (MDP), represented by $(\mathcal{S},\mathcal{A},\mathcal{P}, \mathcal{R},\gamma )$, where $\mathcal{S}$ is the environment's state space, $\mathcal{A}$ is the action space, $\mathcal{P} : \mathcal{S} \times \mathcal{A} \rightarrow  \mathcal{S}$ is the state transition model, $\mathcal{R} : \mathcal{S} \times \mathcal{A} \rightarrow \mathbb{R}$ is the reward function and $\gamma$ is the discount factor of the cumulative reward. Reinforcement learning systems learn the optimal action given the state to maximize a objective defined as a cumulative discounted reward over time. In a typical setting, an agent receives the environment's state and uses it to choose an action based on its policy. In response, the system makes a transition to a new state and provides the reward, and the cycle is repeated. The problem is then to learn an optimal policy for the agent to maximize the total reward over a finite or infinite time horizon in the future.

The key concepts for the notification decision problem are described below.

\textbf{Actions.} $a$ denotes an action in the action space $\mathcal{A}$. We consider a discrete action space consisting of two actions - SEND (send the notification candidate to the user) and  NOT-SEND (the notification candidate is put back in the notification queue for further considerations). Note that every notification candidate has its validity time window ranging from a few hours to a few days. Once a notification candidate reaches its expiry time, it will be either sent or dropped based on its quality. Since this is controlled by an independent logic, it is abstracted out as part of the environment.

\textbf{States.} $s$ denotes a state in the state space $\mathcal{S}$. A state represents a situation in the environment and summarizes all useful historical information. A state $s_t$ at time $t$, has the Markov property, if and only if,
$$P(s_{t+1}| s_1,s_2,..,s_t) = P(s_{t+1}| s_t).$$
Since MDP is the foundation for reinforcement learning algorithms, states must be defined properly to ensure and Markovian system. In our problem setting, we use a plethora of features, including
 \begin{itemize}
\item user's profile features such as locale and network size.
\item dynamic state features such as badge count, number of notifications in the queue, number of notifications sent in the past 24 hours.
\item user's activity features such as user's last visit time, the number of site visits over the past week.
\end{itemize}
In this manner, we allow the users to be part of the environment and represent their interests and context using a rich state representation.

\textbf{Environment.} In standard reinforcement learning, an agent interacts with an environment over a number of discrete time steps. At every time step $t$, the agent receives a state $s_t$ and chooses an action $a_t$. In return, the agent receives the next state $s_{t+1}$ and a scalar reward $r_t$. In our problem, the environment is made up of all users' interests and interactions. A single episode corresponds to a sampled user and their interaction sequence. It consists of all the time steps at which the agent evaluates whether to send the top-ranked notification in the queue over a finite time horizon.

\textbf{Reward.} $r_t$ denotes an immediate reward collected between time $t$ and $t+1$. In this paper, we use a user visit to the platform within the next time step as a reward. The total return $R_t = \sum_{k=t}^\infty \gamma^k r_k$ represents the time-discounted total number of site visits by a user. Here, $\gamma \in (0,1]$ is the discount factor, which controls the trade-offs between the short-term and long-term rewards. The goal of the agent is to maximize this total return to encourage long-term site engagement. The reward can also be defined as notification clicks, or notification disables as negative rewards or a linear combination of them.

\textbf{Policy.} A policy $\pi$ is a mapping from the state space to the action space. In this setting, it makes SEND or NOT-SEND decisions given the state features. A policy can be either deterministic or stochastic. For every Markov decision process, there exists an optimal deterministic policy $\pi^*$, which maximizes the total return from any initial state.

\section{Methodology}
\label{sec:methodology}

We now introduce our offline training framework and offline evaluation method. We also show how setting up a simulated environment can help validate offline training and offline evaluation.

\subsection{Offline Training}
\label{sec:offlineTraining}
One main challenge in applying reinforcement learning to any real-world online recommender systems is the high cost of exploration, which would harm the user experience. That is, if we train a reinforcement learning agent in an online fashion \cite{mnih2013playing}, our users can suffer from an exploratory yet bad policy, which we try to avoid. 

Offline Reinforcement Learning, also known as Batch Reinforcement Learning \cite{lange2012batch} in literature, is a variant of reinforcement learning that the agent learns from a fixed batch of data \cite{levine2020offline}. This variant is suitable for large-scale user-platform interactive applications due to its control over exploration risks, and it naturally fits into existing machine learning infrastructures compared to online reinforcement learning. Our proposed offline solution is a combination of Offline Deep Q-Network (DQN) and data collection with well-controlled online exploration.

Q-learning algorithms are good candidates for offline reinforcement learning attributable to their off-policy nature, that is, they can learn the value of the optimal policy independently of the agent's actions. On the contrary, on-policy learner learns the value of the policy being carried out by the agent, including the exploration steps. While most on-policy algorithms can have their off-policy versions, they require non-trivial importance sampling to adjust the off-policy bias. Importance sampling based estimation can be of very high variance, especially in an offline paradigm.

Among Q-learning algorithms, Deep Q-Network is using Deep Neural Network that takes a state and approximates Q-values for each action based on that state, which has been proved to be successful in tackling high-dimensional state space and gain wide popularity in the recent industry and research advancements \cite{mnih2013playing,van2016deep}. Therefore, we choose Offline Deep Q-Network algorithms for our use-case to learn from the data generated by the complex user behaviors that happened in the real world.

The Q-value function which takes two inputs state ($s$) and action ($a$) under policy $\pi$ is defined as 
\setlength{\arraycolsep}{0.0em}
 \begin{eqnarray}
\label{eq:qFunction}
Q^{\pi}(s_t,a_t)&{ }={ } &E_{\pi}[R_t|s_t=s,a_t=a] \nonumber \\
&{ }={ } &E_{\pi}[{\sum_{k=0}^{\infty}\gamma^k}r_{t+k+1}|s_t=s,a_t=a],
\end{eqnarray}
\setlength{\arraycolsep}{5pt}
 where $\gamma$ is the discount factor, at each time step $t$ the agent with state $s_t$ selects an action $a_t$, observes a reward $r_t$, and $R_t$ is the cumulative long-term reward.

We define the optimal Q-value function
 \begin{equation}
\label{eq:optimalQ}
Q^{{\pi}^*}(s_t, a_t)=\max_{\pi} E_{\pi}(R_t|s_t = s, a_t = a),
\end{equation}
as the maximum expected return achievable across all policies. An optimal policy is easily derived from the optimal $Q^{{\pi}^*}(s, a)$ by selecting the highest-valued action in each state. This optimal $Q^{{\pi}^*}(s, a)$ obeys the Bellman equation:
\begin{equation}
\label{eq:bellman}
Q^{{\pi}^*}(s, a) = E\left(r + \gamma\max_{a^{\prime}} Q^{{\pi}^*}(s^{\prime}, a^{\prime})|s,a \right).
\end{equation}
The Deep Q-Network (DQN) learns a parameterized action-value function
$Q(s, a; \boldsymbol { \theta })$ as a neural network. From Equation \ref{eq:bellman}, the Q-Network can be trained by minimizing the following loss function:
 \begin{equation}
\label{eq:dqnLoss}
\begin{aligned}
L(\boldsymbol { \theta }) &=& E_{s_t,a_t,r_t,s_{t+1}~\mathcal{B}}\left((y_t- Q(s, a; \boldsymbol { \theta }))^2 \right),\\
y_{ t }  &=& r_{ t } + \gamma \max_{ a } Q\left( s_{ t + 1 } , a ; \boldsymbol { \theta } _ { t } \right),
\end{aligned}
\end{equation}
where $\mathcal{B}$ is an offline data batch which contains transition tuples of $\{s_t, a_t, r_t, s_{t+1}\}$. The function approximation $\boldsymbol{\theta}$ can be designed according to characteristics of the application. In this work, we use a fully-connected network, which proved to be sufficient to capture the user interactions. Practically, a few techniques can be employed to improve the learning performance, such as setting up a separate target network \cite{mnih2013playing}, tuning hyper-parameters, using more advanced DQN variants (for example, Double DQN \cite{van2016deep} and dueling DQN \cite{pmlr-v48-wangf16}). Our presented results are based on Double DQN, in which a second network $\boldsymbol { \theta } _ { t } ^ { - }$ is introduced to stabilize the target action-value estimation and reduce the over-estimation well known for the vanilla DQN.  The Double DQN can be trained by minimizing the following loss function:

\setlength{\arraycolsep}{0.0em}
 \begin{eqnarray}
\label{eq:doubleQ}
L(\boldsymbol { \theta }) &{ }={ }& E_{s_t,a_t,r_t,s_{t+1}~\mathcal{B}}\left(( y_ { t } ^ { \mathrm { DoubleQ } }- Q(s, a; \boldsymbol { \theta }))^2 \right),\nonumber\\
 y_ { t } ^ { \mathrm { DoubleQ } } &{ }={ }& r _ { t } + \gamma Q \left( s _ { t + 1 } , \operatorname { argmax } _ { a } Q \left( s _ { t + 1 } , a ; \boldsymbol { \theta } _ { t } \right) , \boldsymbol { \theta } _ { t } ^ { - } \right). \nonumber
\end{eqnarray}
\setlength{\arraycolsep}{5pt}

These DQN variants were originally designed for online off-policy learning \cite{mnih2013playing}. If we only look at the training algorithm, the offline version above is almost the same except that the mini-batch is sampled from an offline data batch instead of an experience buffer in \cite{mnih2013playing}. However, the fundamental difference is that the offline training has no control over the behavior policy. Nor can it explore unseen action trajectories in the batch through interactions with the environment, which is safe for our scenario (to protect user experience). Blending with exploration is outside the scope of the algorithm we are using, as it is both theoretically and practically challenging to guarantee the performance of a learned policy.

Fortunately, we have the control over how the offline data are collected to certain extent, that is, we need to ensure enough exploration in the offline data. We deploy an $\epsilon-$greedy exploration strategy \cite{sandholm1996multiagent} on the baseline policy described in Section \ref{sec:notificationSpacing},

\begin{equation}
    \label{eq:epsilonGreedy}
    \pi^{\epsilon}(s) =  
    \begin{cases}
      \pi_0(s) & \text{with probability } 1-\epsilon,\\
      a \in \text{Unif}(A ) & \text{with probability } \epsilon,
    \end{cases}
\end{equation}
where $\pi_0$ is the baseline policy,  $\text{Unif}(A )$ is a uniform distribution over all possible actions  and $\epsilon$ controls the exploration rate.

Not only is the exploration in data collection important to the error bound of offline learning, it is also critical to the offline policy evaluation, that 
\begin{enumerate}
    \item Data used for offline policy evaluation has to be collected from a stochastic policy with non-zero probability coverage on the state-action space.
    \item The action probabilities have to be correctly recorded.
\end{enumerate}

We will elaborate the discussion in the next section.

\subsection{Offline Evaluation}
\label{sec:offlineEvaluation}
It is crucial to evaluate the performance of the learned policy before risking deployment. Furthermore, we often have more than one algorithm and corresponding hyper-parameter settings, making the offline evaluation an indispensable component of a reinforcement learning training pipeline. The offline evaluation of a reinforcement learning agent requires the estimation of a counterfactual metric of interest from the data collected from an arbitrary but known policy. Model-based approaches for evaluating a reinforcement learning agent from such off-policy data can induce large bias, while the classical importance weighting based non-parametric methods tend to exhibit a high variance for long-term evaluations. We propose a class of importance weighting based methods that can be tuned to obtain a desirable bias-variance trade-off depending on the environment. To this end, we first describe the importance weighting strategy. 

Given $N$ i.i.d. trajectory observations from time $1$ to $T$, $\{(s_{1,i}, a_{1,i}, r_{1,i}, \ldots, s_{T, i}, a_{T, i}, r_{T,i})\}_{i=1}^N$ based on a policy $\pi$ from the joint distribution $p_{\pi}(\cdot)$, we aim to estimate the total expected reward $\theta(\pi^*) = \sum_{t=1}^{T}E_{\pi^*}[r_t]$ corresponding to a target policy $\pi^*$. A estimator of $\theta(\pi^*)$ is given by
\[
\hat{\theta}(\pi^*) = \frac{1}{N} \sum_{i=1}^{N} \sum_{t=1}^{T} r_{t,i}~ w_{t,i}
\]
where $w_{t,i}$ denotes the importance weights adjusted for the mismatch in the distributions under policies $\pi$ and $\pi^*$.

Now it is easy to show that $\hat{\theta}(\pi^*)$ is an unbiased estimator of $\theta(\pi^*)$ for
\begin{align}\label{eq: importance-weight}
w_{t, i} = \frac{p_{\pi^*}(s_{t,i},a_{t,i})}{p_{\pi}(s_{t,i},a_{t,i})},
\end{align}
given that the data generating policy is a stochastic policy $\pi(a \mid s) > 0$ for all $a \in \mathcal{A}$ and $s \in \mathcal{S}$.

In most cases, the functional form of the distributions $p_{\pi}(\cdot)$ and $p_{\pi^*}(\cdot)$ are unknown and hence $w_{t,i}$ needs to be computed/estimated from the data. There are two main obstacles in constructing reliable estimates of $\theta(\pi^*)$ based on $w_{t,i}$: (i) the curse of dimensionality of the state space and (ii) the curse of the horizon. While the former is a well-known problem in supervised learning, the latter is tied to reinforcement learning problems with a long time horizon $T$.
\\~\\
{\bf Action Trajectory Based Weighting \cite{mahmood2014weighted}:} The following estimator avoids the curse of dimensionality of the state space by factorizing $p_{\pi}(\cdot)$ and $p_{\pi^*}(\cdot)$ using the Markov property.
\[
w_{t,i} = \frac{p_{\pi^*}(s_{1,i})~\prod_{j=1}^t \pi^*(a_{j,i}\mid s_{j,i})}{p_{\pi}(s_{1,i})~\prod_{j=1}^t \pi(a_{j,i}\mid s_{j,i})} = \frac{\prod_{j=1}^t \pi^*(a_{j,i}\mid s_{j,i})}{\prod_{j=1}^t \pi(a_{j,i}\mid s_{j,i})},
\]
where the last equality follows from the fact that the distribution of $S_1$ does not depend on the underlying policy. The corresponding estimator of $\theta(\pi^*)$ assign weights to each sample $(s_1,a_1,\ldots,s_t,a_t)$ according to the probability of observing the action trajectory $(a_1,\ldots,a_t)$ under $\pi^*$. Thus, we refer to this method as \emph{Action Trajectory Based Weighting}. Note that if $\pi^*$ is a deterministic policy, i.e.\ $\pi^*(s | a) \in \{0, 1\}$, then this method would assign zero weights to all trajectory that are not feasible under $\pi^*$.

This strategy would suffer from the curse of the horizon, i.e., the variance of the corresponding estimator of $\theta(\pi^*)$ would increase exponentially with the time horizon $T$.
\\~\\
{\bf State Marginalized Weighting \cite{xie2019towards}:} Again using the Markov property, an alternative way of factorizing $w_{t,i}$ in \eqref{eq: importance-weight} is as follows.

\[
w_{t_i} = \frac{p_{\pi^*}(s_{t,i})~ \pi^*(a_{t,i} \mid s_{t,i})}{p_{\pi}(s_{t,i})~ \pi(a_{t,i} \mid s_{t,i})}.
\]
First, we consider the case when the state space is discrete and finite. In this case, $p_{\pi}(s_t)$ can be estimated as
\[
\hat{p}_{\pi}(s_t) = \frac{1}{N} \sum_{i = 1}^n 1_{\{s_{t,i} = s_t\}}.
\]
Next, we estimate $p_{\pi^*}(s_t \mid s_1)$ recursively as follows.

\begin{equation}
\hat{p}_{\pi^*}(s_t) = 
\begin{cases}
       \frac{1}{N} \sum_{i = 1}^n 1_{\{s_{t,i} = s_t\}}  \;\;\;\;\;\;\;\;\;\;\;\;\; \;\;\;\;\;\; \text{if $t=1$};\\
      \frac{1}{N} \sum_{i = 1}^n 1_{\{s_{t,i} = s_t\}} \frac{p_{\pi^*}(s_{t-1,i})~ \pi^*(a_{t-1,i} \mid s_{t-1, i})}{p_{\pi}(s_{t-1,i})~ \pi(a_{t-1,i} \mid s_{t-1, i})} \\ \;\;\;\;\; \;\;\;\;\;\;\;\;\;\;\;\;\;\;\;\;\;\;\;\;\;\;\;\;\;\;\;\;\; \;\;\;\;\;\;\;\;\;\;\;\; \text{if $t>1$}.
    \end{cases}  
    \nonumber 
\end{equation}

This strategy avoids the curse of the horizon by marginalizing over the state dimension. In fact, \cite{xie2019towards} showed that the variance of the corresponding estimator of $\theta(\pi^*)$ is $O(T^3)$. However, the estimations of $p_{\pi}(s_t)$ and $p_{\pi^*}(s_t)$ suffers from the curse of state dimensionality.

In the case of a large state space with multiple features (some/all of which can be continuous), we apply two strategies for dimension reduction:
\begin{enumerate}
\item We remove the state features from offline evaluation that are not influenced by the action (e.g., static features). More precisely, we can work with a reduced state space $h(s_t)$ as long as we have 
\begin{align}\label{eq: dimension reduction}
\frac{p_{\pi^*}(s_t)}{p_{\pi^*}(s_t)} = \frac{p_{\pi^*}(h(s_t))}{p_{\pi^*}(h(s_t))}.
\end{align}
It is to show that following is a sufficient condition for having \eqref{eq: dimension reduction}:
\[
p(s_t \mid h(s_t), a_1,\ldots,a_t) = p(s_t \mid h(s_t)),
\]
which states that all the features that are not in $h(s_t)$ are conditionally independent of the actions given $h(s_t)$. Note that these features (that are not in $h(s_t)$) cannot be removed from model training since they can have an influence on the reward function.

\item We discretize each feature (that is left after the first step) into a fixed number of bins, where the bin size controls the bias-variance trade-off (a larger number of bins would lead to a smaller bias but a larger variance).
\end{enumerate}

{\bf One-Step Correction based Weighting \cite{chen2019top}:} An easy way to construct a biased but low variance estimator is to use the following weights.
\[
w_{t,i} = \frac{\pi^*(s_{t,i} \mid a_{t,i})}{\pi(s_{t,i} \mid a_{t,i})}
\]
This strategy avoids both the curse of dimensionality of the state space and the curse of the long horizon. The corresponding estimator that only corrects for the one-step policy mismatch is no longer unbiased. However, when the variances of both action trajectory based estimator and the state marginalized estimator are too high, the one-step correction might be the only reliable method.

\subsection{Simulation Environment}
\label{sec:simulationEnvironment}
In order to  evaluate  and benchmark offline evaluation methods and offline training algorithms with known ground truth, we build a simplified simulation environment using Open AI GYM \cite{DBLP:journals/corr/BrockmanCPSSTZ16}, to mimic the online notification spacing system. The reason we build such a simulation environment instead of using existing ones in GYM is that learning algorithms and offline evaluation methods that work well in one environment may not work as well in another environment. The closer we build a simulation environment to the real notification environment, the more accurate and useful we get from simulation studies. Section \ref{sec:simulatedExperiments} gives a simulated study using this simulation environment. 

The Markov decision process, which is implemented in the simulator, is described below.
\begin{itemize}
\item \textbf{Environment}: The simulator mimics the notification queue of a user along with the environment. The environment generates notification candidates to be sent to users. We simulate user visits based on two factors - badge count (the number of unseen notifications awaiting the user on the app) and their activeness on the app. We assume that once a user visits the app, the notifications previously sent to them are seen by the user, and hence we reset the badge count for the user.
\item \textbf{Transition process}: The time step for the simulator is set to 4 hours. At each time step, the simulator generates new notifications to arrive in the queue. This is done by sampling a Poisson process, which is scaled by the day and time of the week as well as the demand patterns at that time. Each notification added to the queue consists of a relevance score (used to indicate the rank of the notification in the queue) as well as the time of expiry. Since users also receive near-real-time notifications outside of the queue, their visits and badge counts are affected by those notifications. To get the simulation environment closer to the real-world scenario, we similarly simulate new notifications directly sent to the user outside of the queue and update their badge count as part of the environment. Finally, we check for any expiring notifications available in the queue and remove them from the queue by either sending them to the user with a $50\%$ probability or dropping them altogether.
\item \textbf{State}: We use a simple four-dimensional state comprising of badge count, number of candidate notifications in the queue, time since the start of the week, and user activeness. 
\item \textbf{Actions}: Consistent with the production system, there are two actions, SEND and NOT-SEND.
\item \textbf{Reward}: The numerical reward is 1 if a user visits in the next four hours. We use a reward model, which is a function of user activeness and badge count, which is fit to qualitatively represent actual user performance that is learned from our visit state transition models describe in \cite{yuan2019state}.
\end{itemize}

The simulator roughly captures the notification spacing system, with a time-varying demand generation and user visits to the app. The queuing and de-queuing logic is consistent with the actual notification spacing system. For demand generation and user visits, we qualitatively capture the characteristics of our production models in the simulator.

 Once the offline training algorithm and offline evaluation method is chosen, the offline training of the real-world data is independent and takes no information from this simulation environment. Therefore, any simplification and bias of the simulation environment from the real notification system is not a concern. Such an environment setup can be a low-cost alternative to simulation environments for hybrid learning \cite{shi2019virtual,zou2019reinforcement}.


\section{Experiments}
\label{sec:experiments}
We present both a simulation study and a real-world online experiment to demonstrate the contribution of the proposed framework. 

\subsection{Offline Evaluation in Simulation Environment}
\label{sec:simulatedExperiments}
\begin{figure}[h]
\center
\includegraphics[width=0.75\linewidth]{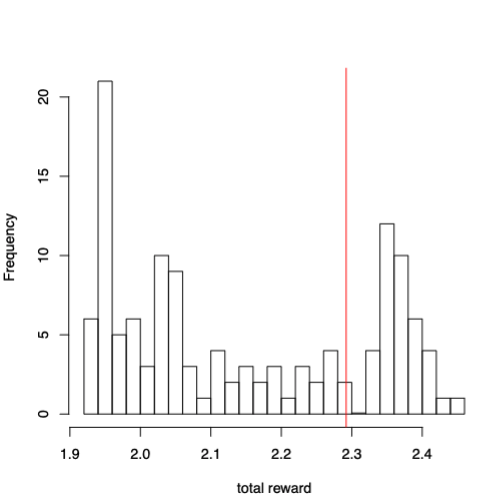}
\caption{\label{fig:hyperParameter} Distribution of policy performance with respect to the baseline in red } 
\end{figure}

First, we show why offline evaluation is crucial to the offline reinforcement learning framework. Without true interaction with the environment, offline training can be unstable in terms of convergence and policy performance. We generated both training and validation data using an exploration policy describe in Equation \ref{eq:epsilonGreedy} in the notification simulation environment described in Section \ref{sec:simulationEnvironment}. We then trained 128 Double DQN policies using the same training data under various hyper parameters (batch size, learning rate, number of batches with a fixed target network, network layer size, etc.). Figure \ref{fig:hyperParameter} shows the online evaluation distribution in the simulation environment of the 128 policies, which can be regarded as the ground truth of the policy performance. The red vertical line is the performance of the behavior exploration policy. Only 38 out of 128 policies delivered total rewards exceeding that of the behavior policy. This demonstrates that in the offline learning setting, the learned policy is not guaranteed to perform better than the behavior policy and that hyper-parameter tuning and offline evaluation is necessary. For real-world applications, online evaluation can be risky and expensive, and offline evaluation is desired.

Next, we show how the proposed state-marginalized importance sample provides better bias-variance trade-offs compared with trajectory matching and one-step matching described in Section \ref{sec:offlineEvaluation}. We evaluate the same 128 policies above using the three offline evaluation methods, where we used the self-normalized version of importance weights, i.e.\ $w_{t, i} / (\frac{1}{N} \sum_{i=1}^N w_{t,i})$, which is known to be more stable in practice.

\begin{figure}[h]
\centering
    \includegraphics[width=0.75\linewidth]{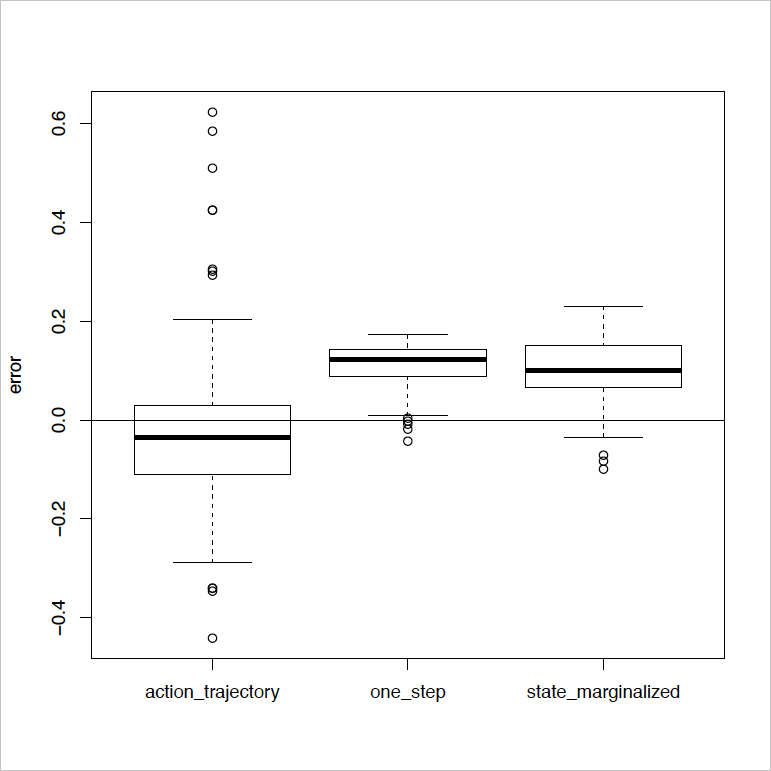} 
\caption{\label{fig:simulator} Bias-variance comparison between three methods} 
\end{figure}

Figure \ref{fig:simulator} shows the box plot of the error between offline and online estimates of the cumulative reward across all learned policies, obtained from one-step, action-trajectory  and state-marginalized importance weighting techniques. We observe that action trajectory based importance weighting technique provides the lowest biased estimate of the cumulative reward but has the highest variance. One step importance weighting technique provides the highest biased estimate of the cumulative reward but has the lowest variance. The state-marginalized importance weighting estimate has a better balance between bias and variance of the estimate. In this simulation, we chose the number of bins to be 10 for discretizing the states. The bias of the estimate from this technique can be lowered by increasing the number of bins but at the expense of higher variance. However, the variance can be easily estimated in the production-setting, thereby allowing us to optimally choose the number of bins to control bias-variance trade-offs in our estimation.

\begin{figure*}[ht]
\centering
  \includegraphics[width = 0.9\linewidth]{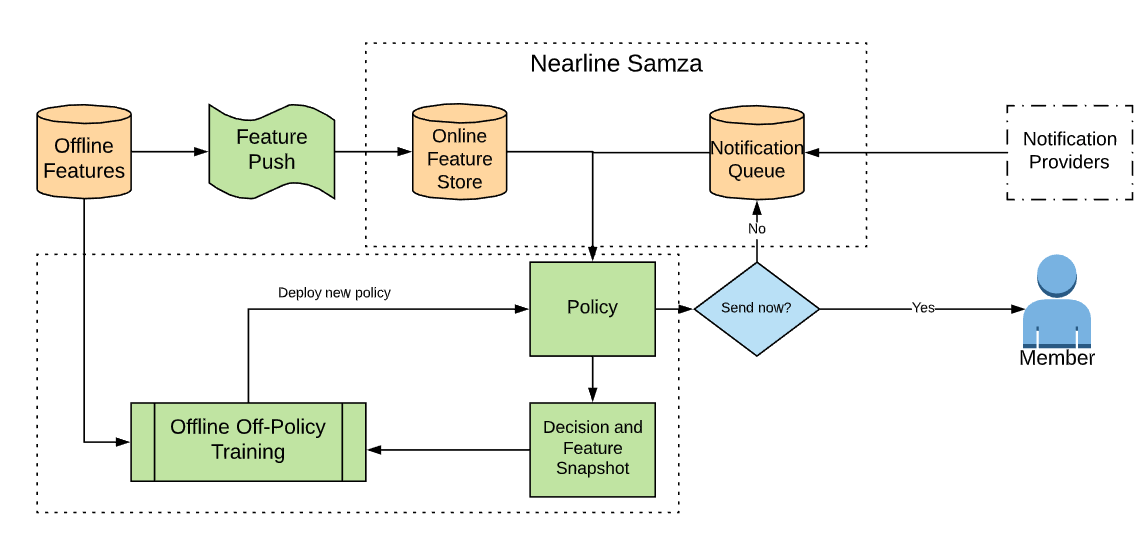}
  \caption{System architecture}
  \label{fig:infra}
\end{figure*}

\subsection{Online Experiments in Notification Spacing}
\label{sec:onlineExperiments}
 Figure \ref{fig:infra} illustrates the system architecture of this application. All policies, regardless of supervised or reinforcement learning, are served in a nearline Samza service \cite{noghabi2017samza}. Various notification providers send notification candidates to the Samza service and are queued in its data store for future evaluations. Every user's notification queue is evaluated by a policy every few hours for a send-or-not decision for its top scored notification. A policy takes online features, makes decisions, and then snapshots its decisions and features to a Hadoop Distributed File System (HDFS) \cite{shvachko2010hadoop}. Additional offline features can be used for offline training and pushed to Samza data stores for nearline serving.

Under this architecture, we collected one-week snapshot data from a small percentage of LinkedIn users using an epsilon-greedy behavior policy $\pi^{\epsilon}$ on top of the baseline policy described in \ref{sec:notificationSpacing}. The data was then joined with other log data to construct the tuples $(s_t,a_t,s_{t+1},r_{t})$ for offline training. The state features we use are the same as the features in the baseline supervised model to allow a fair comparison. Important state features include the app badge count, the user's last visit time, the number of notifications received over the past week and other user profile features. We then train the Double DQN models described in Section \ref{sec:offlineTraining} using a fully-connected 3-layer neural network with different hyper-parameters (number of inner nodes, learning rate, batch size, number of batches between target network updates, etc.). We apply the state marginalized weighting to estimate the expected total return of the learned policies. Given the large continuous state space, we apply  the dimension reduction and discretization strategies described in Section \ref{sec:offlineEvaluation} for the state marginalized weighting to further reduce the estimation variance. The top performer (or performers) selected by offline evaluation was deployed in the Samza service on a certain percentage of total users for an online A/B test compared with the baseline policy.  We are interested in user engagement and notification interactions, which can be characterized by the following metrics.

\begin{itemize}
\item \emph{Sessions}: A session is a collection of full-page views made by a single user on the same device type. Two sessions are separated by $30$ minutes of zero activity.  This is a widely used metric on user engagement across social networks.
\item \emph{Notification  cards}: the total number of notification cards served to users after removing potential duplicates. This is a metric capturing notification volume. 
\item \emph{Notification  CTR}: This metric measures the average click-through-rate of notifications sent to a user in a day. This is a metric capturing notification interactions. 
\item \emph{ Notification unfollow total}: the total number of notification unfollow actions taken by users. This is negative feedback from users. 
\end{itemize}

\begin{table}[!h]
\caption{Online A/B results for delivery time optimization}
\centering
  \begin{tabular}{ | l | c | }
    \hline
    Metric & vs. Baseline policy  \\
    \hline
    \hline
   Sessions & +0.30\% \\
     \hline
    Notification  cards  & -3.49\%\\
    \hline
    Notification  CTR & +4.53\% \\
    \hline
    Notification  unfollow total & -4.37\% \\
    
    \hline
  \end{tabular}
\label{tab:abtest}
\end{table}

Table \ref{tab:abtest} shows the full-week A/B test results, and the numbers are all statistically significant (with p-value $<0.05$). Compared with the baseline policy, the new policy from offline reinforcement learning increased the total sessions by $0.3\%$, which is considered a moderate gain in a volume neutral iteration, but very impressive given that the total notification volume is reduced by $3.49\%$. The $4.53\%$ increase in notification CTR and $4.37\%$ decrease in notification unfollow total are mainly driven by the reduction in notification volume. These numbers demonstrate the business impact of our proposed framework and suggest that notifications were delivered at better timing and lower frequency, suggesting more optimality over the supervised approach.

In addition to business impact, we would like to point out that the offline reinforcement learning training and evaluation framework  provides a significant increase in iteration speed in our ecosystem. At LinkedIn, we constantly test new features and new learning algorithms, which requires a modeling iteration cycle consisting of offline training and evaluation, online tuning (if needed) before we take it to the online A/B test to conclude whether the new model is better than its baseline. Therefore,  we measure the iteration speed by the time it takes for each steps. Although reinforcement learning algorithms in general take more computation resources and longer time to converge than most supervised learning models, the difference in training and evaluation (in hours) is negligible when we take the online tuning into consideration.   As described in Section \ref{sec:notificationSpacing}, $\tau$  has to be tuned online for a trade-off between short-term and long-term engagement. In fact, many supervised model-based frameworks typically resort to online tuning like the approach used in \cite{talosTutorial}, as there is usually a gap between the model predictions and the service decisions/actions.  The tuning typically takes 1-3 weeks for notifications as site engagement responses takes days to show up. This leads to tedious and time-consuming efforts for many practitioners. In contrast, the offline evaluation system, deployed as a part of the reinforcement learning model, does not require an online tuning cycle after the policy selection done in the offline evaluation. The selected policy can go directly to the online A/B test to draw a conclusion on a modeling iteration.


\section{Discussion}
\label{sec:conclusion} 

In this paper, we propose an offline reinforcement learning framework covering data collection, offline learning, offline evaluation, assisting simulation environment, and other practical considerations. We argue and demonstrate the benefits of such a framework  as a more principled paradigm to optimize notification decisions over supervised learning approaches. Practically, it shortens the  modeling iteration cycle for real-world systems that constantly improve with new features and retraining. 

One of the limitations of our presented results is that we trained and tested this framework in a one-week frame. It would be interesting to see whether the framework can improve even longer-term engagement. This is limited by the measurement cost of truly long-term engagement, say a one-year scope. 

Based on its positive impact and iteration speed gains, the reinforcement learning application in Section \ref{sec:onlineExperiments} was fully deployed at LinkedIn. We hope our work could motivate wider adoption of reinforcement learning for the mainstream recommender systems. We will continue to work on how to learn the optimal policy more efficiently offline and how to evaluate more accurately offline.

\section*{Acknowledgment}

We are thankful to Shaunak Chatterjee, Yan Gao, Cyrus DiCiccio, Bee-Chung Chen, Deepak Agawal, Matthew Walker,   Romer rosales, Shipeng Yu and Mohsen Jamali for their detailed and insightful feedback during the development of this work.

\bibliographystyle{IEEEtran}
\bibliography{notification_rl}

\begin{thebibliography}{10}
\providecommand{\url}[1]{#1}
\csname url@samestyle\endcsname
\providecommand{\newblock}{\relax}
\providecommand{\bibinfo}[2]{#2}
\providecommand{\BIBentrySTDinterwordspacing}{\spaceskip=0pt\relax}
\providecommand{\BIBentryALTinterwordstretchfactor}{4}
\providecommand{\BIBentryALTinterwordspacing}{\spaceskip=\fontdimen2\font plus
\BIBentryALTinterwordstretchfactor\fontdimen3\font minus
  \fontdimen4\font\relax}
\providecommand{\BIBforeignlanguage}[2]{{%
\expandafter\ifx\csname l@#1\endcsname\relax
\typeout{** WARNING: IEEEtran.bst: No hyphenation pattern has been}%
\typeout{** loaded for the language `#1'. Using the pattern for}%
\typeout{** the default language instead.}%
\else
\language=\csname l@#1\endcsname
\fi
#2}}
\providecommand{\BIBdecl}{\relax}
\BIBdecl

\bibitem{pielot2014situ}
M.~Pielot, K.~Church, and R.~De~Oliveira, ``An in-situ study of mobile phone
  notifications,'' in \emph{Proceedings of the 16th international conference on
  Human-computer interaction with mobile devices \& services}.\hskip 1em plus
  0.5em minus 0.4em\relax ACM, 2014, pp. 233--242.

\bibitem{okoshi2019real}
T.~Okoshi, K.~Tsubouchi, and H.~Tokuda, ``Real-world product deployment of
  adaptive push notification scheduling on smartphones,'' in \emph{Proceedings
  of the 25th ACM SIGKDD International Conference on Knowledge Discovery \&
  Data Mining}, 2019, pp. 2792--2800.

\bibitem{gupta2016email}
R.~Gupta, G.~Liang, H.-P. Tseng, R.~K. Holur~Vijay, X.~Chen, and R.~Rosales,
  ``Email volume optimization at linkedin,'' in \emph{Proceedings of the 22nd
  ACM SIGKDD International Conference on Knowledge Discovery and Data
  Mining}.\hskip 1em plus 0.5em minus 0.4em\relax ACM, 2016, pp. 97--106.

\bibitem{gupta2017optimizing}
R.~Gupta, G.~Liang, and R.~Rosales, ``Optimizing email volume for sitewide
  engagement,'' in \emph{Proceedings of the 2017 ACM on Conference on
  Information and Knowledge Management}.\hskip 1em plus 0.5em minus 0.4em\relax
  ACM, 2017, pp. 1947--1955.

\bibitem{yuan2019state}
Y.~Yuan, J.~Zhang, S.~Chatterjee, S.~Yu, and R.~Rosales, ``A state transition
  model for mobile notifications via survival analysis,'' in \emph{Proceedings
  of the Twelfth ACM International Conference on Web Search and Data Mining},
  2019, pp. 123--131.

\bibitem{gao2018near}
Y.~Gao, V.~Gupta, J.~Yan, C.~Shi, Z.~Tao, P.~Xiao, C.~Wang, S.~Yu, R.~Rosales,
  A.~Muralidharan \emph{et~al.}, ``Near real-time optimization of
  activity-based notifications,'' in \emph{Proceedings of the 24th ACM SIGKDD
  International Conference on Knowledge Discovery \& Data Mining}, 2018, pp.
  283--292.

\bibitem{zhao2018notification}
B.~Zhao, K.~Narita, B.~Orten, and J.~Egan, ``Notification volume control and
  optimization system at pinterest,'' in \emph{Proceedings of the 24th ACM
  SIGKDD International Conference on Knowledge Discovery \& Data Mining}, 2018,
  pp. 1012--1020.

\bibitem{agarwal2011click}
D.~Agarwal, B.-C. Chen, P.~Elango, and X.~Wang, ``Click shaping to optimize
  multiple objectives,'' in \emph{Proceedings of the 17th ACM SIGKDD
  international conference on Knowledge discovery and data mining}.\hskip 1em
  plus 0.5em minus 0.4em\relax ACM, 2011, pp. 132--140.

\bibitem{talosTutorial}
\BIBentryALTinterwordspacing
K.~Basu, C.~DiCiccio, B.~Gavin, V.~Gupta, and Y.~Ouyang, ``Bayesian
  optimization for balancing metrics in recommender systems,''
  \emph{International Joint Conference on Artificial Intelligence}, 2020.
  [Online]. Available:
  \url{https://sites.google.com/view/ijcai2020-linkedin-bayesopt/home}
\BIBentrySTDinterwordspacing

\bibitem{levine2020offline}
S.~Levine, A.~Kumar, G.~Tucker, and J.~Fu, ``Offline reinforcement learning:
  Tutorial, review, and perspectives on open problems,'' \emph{arXiv preprint
  arXiv:2005.01643}, 2020.

\bibitem{lange2012batch}
S.~Lange, T.~Gabel, and M.~Riedmiller, ``Batch reinforcement learning,'' in
  \emph{Reinforcement learning}.\hskip 1em plus 0.5em minus 0.4em\relax
  Springer, 2012, pp. 45--73.

\bibitem{agarwal2019striving}
R.~Agarwal, D.~Schuurmans, and M.~Norouzi, ``Striving for simplicity in
  off-policy deep reinforcement learning,'' 2019.

\bibitem{chen2019top}
M.~Chen, A.~Beutel, P.~Covington, S.~Jain, F.~Belletti, and E.~H. Chi, ``Top-k
  off-policy correction for a reinforce recommender system,'' in
  \emph{Proceedings of the Twelfth ACM International Conference on Web Search
  and Data Mining}, 2019, pp. 456--464.

\bibitem{ie2019reinforcement}
E.~Ie, V.~Jain, J.~Wang, S.~Narvekar, R.~Agarwal, R.~Wu, H.-T. Cheng,
  M.~Lustman, V.~Gatto, P.~Covington \emph{et~al.}, ``Reinforcement learning
  for slate-based recommender systems: A tractable decomposition and practical
  methodology,'' \emph{arXiv preprint arXiv:1905.12767}, 2019.

\bibitem{fujimoto2019off}
S.~Fujimoto, D.~Meger, and D.~Precup, ``Off-policy deep reinforcement learning
  without exploration,'' in \emph{International Conference on Machine
  Learning}.\hskip 1em plus 0.5em minus 0.4em\relax PMLR, 2019, pp. 2052--2062.

\bibitem{kumar2020conservative}
A.~Kumar, A.~Zhou, G.~Tucker, and S.~Levine, ``Conservative q-learning for
  offline reinforcement learning,'' \emph{arXiv preprint arXiv:2006.04779},
  2020.

\bibitem{mahmood2014weighted}
A.~R. Mahmood, H.~Van~Hasselt, and R.~S. Sutton, ``Weighted importance sampling
  for off-policy learning with linear function approximation.'' in \emph{NIPS},
  2014, pp. 3014--3022.

\bibitem{jiang2016doubly}
N.~Jiang and L.~Li, ``Doubly robust off-policy value evaluation for
  reinforcement learning,'' in \emph{International Conference on Machine
  Learning}.\hskip 1em plus 0.5em minus 0.4em\relax PMLR, 2016, pp. 652--661.

\bibitem{thomas2016data}
P.~Thomas and E.~Brunskill, ``Data-efficient off-policy policy evaluation for
  reinforcement learning,'' in \emph{International Conference on Machine
  Learning}.\hskip 1em plus 0.5em minus 0.4em\relax PMLR, 2016, pp. 2139--2148.

\bibitem{xie2019towards}
T.~Xie, Y.~Ma, and Y.-X. Wang, ``Towards optimal off-policy evaluation for
  reinforcement learning with marginalized importance sampling,'' \emph{arXiv
  preprint arXiv:1906.03393}, 2019.

\bibitem{sutton1998reinforcement}
R.~S. Sutton and A.~G. Barto, \emph{Reinforcement learning: An
  introduction}.\hskip 1em plus 0.5em minus 0.4em\relax MIT press Cambridge.,
  1998.

\bibitem{van2018deep}
H.~Van~Hasselt, Y.~Doron, F.~Strub, M.~Hessel, N.~Sonnerat, and J.~Modayil,
  ``Deep reinforcement learning and the deadly triad,'' \emph{arXiv preprint
  arXiv:1812.02648}, 2018.

\bibitem{mehrotra2015designing}
A.~Mehrotra, M.~Musolesi, R.~Hendley, and V.~Pejovic, ``Designing
  content-driven intelligent notification mechanisms for mobile applications,''
  in \emph{Proceedings of the 2015 ACM International Joint Conference on
  Pervasive and Ubiquitous Computing}.\hskip 1em plus 0.5em minus 0.4em\relax
  ACM, 2015, pp. 813--824.

\bibitem{pielot2014didn}
M.~Pielot, R.~de~Oliveira, H.~Kwak, and N.~Oliver, ``Didn't you see my
  message?: predicting attentiveness to mobile instant messages,'' in
  \emph{Proceedings of the SIGCHI Conference on Human Factors in Computing
  Systems}.\hskip 1em plus 0.5em minus 0.4em\relax ACM, 2014, pp. 3319--3328.

\bibitem{pielot2017beyond}
M.~Pielot, B.~Cardoso, K.~Katevas, J.~Serr{\`a}, A.~Matic, and N.~Oliver,
  ``Beyond interruptibility: Predicting opportune moments to engage mobile
  phone users,'' \emph{Proceedings of the ACM on Interactive, Mobile, Wearable
  and Ubiquitous Technologies}, vol.~1, no.~3, pp. 1--25, 2017.

\bibitem{wu2017returning}
Q.~Wu, H.~Wang, L.~Hong, and Y.~Shi, ``Returning is believing: Optimizing
  long-term user engagement in recommender systems,'' in \emph{Proceedings of
  the 2017 ACM on Conference on Information and Knowledge Management}, 2017,
  pp. 1927--1936.

\bibitem{okoshi2015attelia}
T.~Okoshi, J.~Ramos, H.~Nozaki, J.~Nakazawa, A.~K. Dey, and H.~Tokuda,
  ``Attelia: Reducing user's cognitive load due to interruptive notifications
  on smart phones,'' in \emph{2015 IEEE International Conference on Pervasive
  Computing and Communications (PerCom)}.\hskip 1em plus 0.5em minus
  0.4em\relax IEEE, 2015, pp. 96--104.

\bibitem{okoshi2015reducing}
------, ``Reducing users' perceived mental effort due to interruptive
  notifications in multi-device mobile environments,'' in \emph{Proceedings of
  the 2015 ACM International Joint Conference on Pervasive and Ubiquitous
  Computing}, 2015, pp. 475--486.

\bibitem{pejovic2014interruptme}
V.~Pejovic and M.~Musolesi, ``Interruptme: designing intelligent prompting
  mechanisms for pervasive applications,'' in \emph{Proceedings of the 2014 ACM
  International Joint Conference on Pervasive and Ubiquitous Computing}, 2014,
  pp. 897--908.

\bibitem{pielot2015attention}
M.~Pielot, T.~Dingler, J.~S. Pedro, and N.~Oliver, ``When attention is not
  scarce-detecting boredom from mobile phone usage,'' in \emph{Proceedings of
  the 2015 ACM international joint conference on pervasive and ubiquitous
  computing}, 2015, pp. 825--836.

\bibitem{okoshi2017attention}
T.~Okoshi, K.~Tsubouchi, M.~Taji, T.~Ichikawa, and H.~Tokuda, ``Attention and
  engagement-awareness in the wild: A large-scale study with adaptive
  notifications,'' in \emph{2017 ieee international conference on pervasive
  computing and communications (percom)}.\hskip 1em plus 0.5em minus
  0.4em\relax IEEE, 2017, pp. 100--110.

\bibitem{zou2019reinforcement}
L.~Zou, L.~Xia, Z.~Ding, J.~Song, W.~Liu, and D.~Yin, ``Reinforcement learning
  to optimize long-term user engagement in recommender systems,'' in
  \emph{Proceedings of the 25th ACM SIGKDD International Conference on
  Knowledge Discovery \& Data Mining}, 2019, pp. 2810--2818.

\bibitem{zhao2018recommendations}
X.~Zhao, L.~Zhang, Z.~Ding, L.~Xia, J.~Tang, and D.~Yin, ``Recommendations with
  negative feedback via pairwise deep reinforcement learning,'' in
  \emph{Proceedings of the 24th ACM SIGKDD International Conference on
  Knowledge Discovery \& Data Mining}, 2018, pp. 1040--1048.

\bibitem{hughes2019generating}
J.~W. Hughes, K.-h. Chang, and R.~Zhang, ``Generating better search engine text
  advertisements with deep reinforcement learning,'' in \emph{Proceedings of
  the 25th ACM SIGKDD International Conference on Knowledge Discovery \& Data
  Mining}, 2019, pp. 2269--2277.

\bibitem{wang2020incremental}
P.~Wang, K.~Liu, L.~Jiang, X.~Li, and Y.~Fu, ``Incremental mobile user
  profiling: Reinforcement learning with spatial knowledge graph for modeling
  event streams,'' in \emph{Proceedings of the 26th ACM SIGKDD International
  Conference on Knowledge Discovery \& Data Mining}, 2020, pp. 853--861.

\bibitem{swaminathan2016off}
A.~Swaminathan, A.~Krishnamurthy, A.~Agarwal, M.~Dud{\'\i}k, J.~Langford,
  D.~Jose, and I.~Zitouni, ``Off-policy evaluation for slate recommendation,''
  \emph{arXiv preprint arXiv:1605.04812}, 2016.

\bibitem{mnih2013playing}
V.~Mnih, K.~Kavukcuoglu, D.~Silver, A.~Graves, I.~Antonoglou, D.~Wierstra, and
  M.~Riedmiller, ``Playing atari with deep reinforcement learning,''
  \emph{arXiv preprint arXiv:1312.5602}, 2013.

\bibitem{van2016deep}
H.~Van~Hasselt, A.~Guez, and D.~Silver, ``Deep reinforcement learning with
  double q-learning,'' in \emph{Proceedings of the AAAI Conference on
  Artificial Intelligence}, vol.~30, no.~1, 2016.

\bibitem{pmlr-v48-wangf16}
Z.~Wang, T.~Schaul, M.~Hessel, H.~Hasselt, M.~Lanctot, and N.~Freitas,
  ``Dueling network architectures for deep reinforcement learning,'' in
  \emph{Proceedings of The 33rd International Conference on Machine Learning},
  2016, pp. 1995--2003.

\bibitem{sandholm1996multiagent}
T.~W. Sandholm and R.~H. Crites, ``Multiagent reinforcement learning in the
  iterated prisoner's dilemma,'' \emph{Biosystems}, vol.~37, no. 1-2, pp.
  147--166, 1996.

\bibitem{DBLP:journals/corr/BrockmanCPSSTZ16}
\BIBentryALTinterwordspacing
G.~Brockman, V.~Cheung, L.~Pettersson, J.~Schneider, J.~Schulman, J.~Tang, and
  W.~Zaremba, ``Openai gym,'' \emph{CoRR}, vol. abs/1606.01540, 2016. [Online].
  Available: \url{http://arxiv.org/abs/1606.01540}
\BIBentrySTDinterwordspacing

\bibitem{shi2019virtual}
J.-C. Shi, Y.~Yu, Q.~Da, S.-Y. Chen, and A.-X. Zeng, ``Virtual-taobao:
  Virtualizing real-world online retail environment for reinforcement
  learning,'' in \emph{Proceedings of the AAAI Conference on Artificial
  Intelligence}, vol.~33, no.~01, 2019, pp. 4902--4909.

\bibitem{noghabi2017samza}
S.~A. Noghabi, K.~Paramasivam, Y.~Pan, N.~Ramesh, J.~Bringhurst, I.~Gupta, and
  R.~H. Campbell, ``Samza: stateful scalable stream processing at linkedin,''
  \emph{Proceedings of the VLDB Endowment}, vol.~10, no.~12, pp. 1634--1645,
  2017.

\bibitem{shvachko2010hadoop}
K.~Shvachko, H.~Kuang, S.~Radia, and R.~Chansler, ``The hadoop distributed file
  system,'' in \emph{2010 IEEE 26th symposium on mass storage systems and
  technologies (MSST)}.\hskip 1em plus 0.5em minus 0.4em\relax Ieee, 2010, pp.
  1--10.

\end{thebibliography}

\end{document}